\documentclass{article}
\usepackage{spconf,amsmath,graphicx}

\usepackage{amsmath,amsfonts,bm}

\def\eqref#1{equation~\ref{#1}}

\def\1{\bm{1}}

\DeclareMathAlphabet{\mathsfit}{\encodingdefault}{\sfdefault}{m}{sl}
\SetMathAlphabet{\mathsfit}{bold}{\encodingdefault}{\sfdefault}{bx}{n}

\DeclareMathOperator*{\argmax}{arg\,max}
\DeclareMathOperator*{\argmin}{arg\,min}

\title{Beta quantile regression for Robust estimation of uncertainty in the presence of outliers}
\author{Author 1 \and Author 2\footnotemark[1] \and Author 3}
\name{Haleh Akrami\thanks{*These authors contributed equally to this work.}*$^{1}$, Omar Zamzam*$^{1}$, Anand Joshi$^{1}$, Sergul Aydore$^{2}$,  Richard Leahy$^{1}$}
\address{$^{1}$Department of Electrical Engineering, University of Southern California, USA \\ 
$^{2}$ Amazon Web Services, New York, USA \\}

\begin{document}
%
\maketitle
\begin{abstract}

Quantile Regression (QR) can be used to estimate aleatoric uncertainty in deep neural networks and can generate prediction intervals. Quantifying uncertainty is particularly important in critical applications such as clinical diagnosis, where a realistic assessment of uncertainty is essential in determining disease status and planning the appropriate treatment. The most common application of quantile regression models is in cases where the parametric likelihood cannot be specified. Although quantile regression is quite robust to outlier response observations, it can be sensitive to outlier covariate observations (features). Outlier features can compromise the performance of deep learning regression problems such as style translation, image reconstruction, and deep anomaly detection, potentially leading to misleading conclusions. To address this problem, we propose a robust solution for quantile regression that incorporates concepts from robust divergence. We compare the performance of our proposed method with (i) least trimmed quantile regression and (ii) robust regression based on the regularization of case-specific parameters in a simple real dataset in the presence of outlier. These methods have not been applied in a deep learning framework. We also demonstrate the applicability of the proposed method by applying it to a medical imaging translation task using diffusion models.
\end{abstract}
\section{Introduction}
Quantile regression offers an alternative to mean regression in various applications where accurate predictions and their associated reliability are crucial. For instance, in clinical diagnosis, a realistic assessment of prediction uncertainty is essential for determining disease status and planning appropriate treatment. In the context of deep learning, two types of uncertainties are encountered: aleatoric and epistemic. Aleatoric uncertainty arises from the inherent stochasticity of the data, while epistemic uncertainty — often referred to as model uncertainty — is due to limitations in the model itself. It's worth noting that an infinite amount of training data would not reduce aleatoric uncertainty, although it could mitigate epistemic uncertainty. A multitude of methods exist for estimating these uncertainties, including Gaussian process regression, uncertainty-aware neural networks, Bayesian neural networks, and ensemble methods\cite{skafte2019reliable,romano2019conformalized, gawlikowski2023survey}.

Recent studies have proposed to use conditional quantile regression to estimate aleatoric uncertainty in neural networks  \cite{romano2019conformalized,tagasovska2019single,akrami2021quantile, akrami2021deep,angelopoulos2022image} and showed that it can compute well-calibrated intervals. The most common application of quantile regression models is in cases where parametric likelihood cannot be specified \cite{yu2001bayesian}. Similar to the classical regression analysis which estimates the conditional mean, the $\alpha$-th quantile regression $(0<\alpha<1)$ seeks a solution to the following minimization problem
\cite{yu2001bayesian}:
\begin{align}
\argmin \limits_{\theta}\sum_{i}\rho_{\alpha}(y_{i}-f_{\theta}(x_{i})),
\label{q_loss}
\end{align}
where $x_i$ are the inputs, $y_i$ are the responses, $f$ is the model paramaterized by $\theta$, and  $\rho_{\alpha}$ is the \emph{check function} or \emph{pinball loss} \cite{yu2001bayesian} defined as: \[
\rho_{\alpha}(y_{i}-f_{\theta}(x_{i})) =
\begin{cases}
    (y_{i}-f_{\theta}(x_{i}))\alpha, & \text{if } y_{i}\geq f_{\theta}(x_{i})\\
    (f_{\theta}(x_{i}) - y_{i})(1-\alpha), & \text{if } y_{i}<f_{\theta}(x_{i})
\end{cases}
\]

It has been shown that minimization of the loss function in (\ref{q_loss})  is  equivalent to maximizing the likelihood function formed by combining independently distributed asymmetric Laplace densities \cite{yu2001bayesian},
\begin{align*}
\argmax \limits_{\theta} L(\theta)=\frac{\alpha(1-\alpha)}{\sigma} \exp\left\{\frac{-\sum_i \rho_{\alpha}(y_{i}-f_{\theta}(x_{i}))}{\sigma}\right\}.
\end{align*}
where $\alpha$ is the quantile and $\sigma$ is the scale parameter. 

Recently, quantile regression has been employed for uncertainty estimation in regression tasks such as image translation \cite{akrami2022deep} and anomaly detection \cite{angelopoulos2022image} in medical imaging. In these domains, obtaining a reliable uncertainty estimate is of critical importance. Compared to alternative methods for uncertainty estimation, such as sampling using generative models \cite{angelopoulos2022image} or Bayesian uncertainty estimation, quantile regression offers computational efficiency and speed, and does not require sampling.
 
 Statistical machine learning models that involve maximizing likelihood are particularly sensitive to outliers \cite{huber2011robust}. 
Although quantile regression is quite robust to outlying response observations, it can be sensitive to outlier covariate observations (features). It has been shown that perturbing a single ($x_i$, $y_i$) data point in an arbitrary manner can force all quantile regression hyperplanes to intersect at the perturbed point \cite{neykov2012least}. Despite that, it's important to highlight that only a limited number of papers have explored the robustness of quantile regression in the context of covariate observations, particularly within deep learning frameworks.

We outline our contributions in this paper as follows:
(i) We propose a robust quantile regression approach that leverages concepts from robust divergence.
(ii) We compare the performance of our proposed method, particularly in the presence of outliers, to existing techniques such as Least Trimmed Quantile Regression \cite{neykov2012least}, which serves as the only available baseline, and robust regression methods that rely on the regularization of case-specific parameters, in both a simple dataset and a simulated dataset.
(iii) Finally, to illustrate the practical utility of the proposed method, we apply it to a medical imaging translation task, employing state-of-the-art diffusion models. 

\section{Method}

We start by briefly explaining the formulation of least-trimmed quantile regression \cite{neykov2012least} and robust regression based on the regularization of case-specific parameters.

\subsection{Least Trim Quantile Regression (TQR)}
The objective function for TQR is defined as:
\begin{align}
\argmin \limits_{\theta}\sum_{I_{C}}\rho_{\alpha}(y_{i}-f_{\theta}(x_{i}))
\label{q_loss2}
\end{align}
Where $I_{C}$ is the subset of samples with C samples from the training dataset that generates the smallest error. The optimization is similar to quantile regression with an additional iterative process. After initializing with C random samples, at each iteration, the samples with the smallest error are chosen  for training in the next iteration and the process is repeated until there are not any significant changes in loss value comparing to that of the previous iteration. We utilized TQR within a gradient descent optimization framework, where we used only the subset of the batch with the lowest error for backpropagation. 

\subsection{Robust regression based on regularization of case-specific parameters (RCP)}

She and Owen \cite{she2011outlier} proposed a robust regression method using the case-specific indicators in a mean shift model with the regularization method. By generalizing their method to quantile regression, the final loss can be simplified to:
\begin{align}
\argmin \limits_{\theta}\sum_{i}\rho_{\alpha}(y_{i}-f_{\theta}(x_{i})-\gamma_{i})+\lambda \sum_{i} |\gamma_{i}|
\label{q2_loss}
\end{align}
This optimization can be solved using an alternative approach and soft margin thresholding. RCP can be used for any
likelihood based model.

\subsection{$\beta$-quantile regression ($\beta$-QR)}
For parameter estimation, maximizing the likelihood is equivalent to minimizing the KL-divergence between the empirical distribution of the input and statistical model $q(\phi)$. Similarly  a robust $\beta$-loss ($L_{\beta}$) can be derived by replacing the KL-divergence with the $\beta$-divergence $D_{\beta}$ \cite{basu1998robust,akrami2022robust, akrami2020robust}.
\begin{align*}
D_{\beta} (f(x) || g(x)) &=\frac{1}{\beta}\int
(f(x)^{\beta}g(x)^{\beta})f(x)dx\\
&-
\frac{1}{\beta+1} \int(f(x)^{\beta+1}  g(x)^{\beta+1})dx
\end{align*}

\begin{align*}
L_{\beta}=\frac{1}{N}\sum \frac{\exp(\beta l(x_{i}, q(\phi)))-1}{\beta}+ \frac{1}{\beta+1}\int q(\phi)^{\beta+1}
\end{align*}
where $l(x_{i}, q(\phi))$ denotes the log-likelihood of observation $x_{i}$. This loss assigns a weight to each observation based on the likelihood's magnitude, mitigating the influence of outliers on model training \cite{basu1998robust}. In the case of quantile regression, the loss can be simplified to:
\begin{align}
 L_{\beta \alpha}=\frac{1}{N}\sum \frac{\exp(-\beta \rho_{\alpha}((y_{i}-f_{\theta}(x_{i}))/\sigma)-1}{\beta}
 \label{robust_loss}\end{align}
The hyperparameter $\sigma$ can be assumed to be 1 for simplicity. This loss can be interpreted as an M-estimate. The hyperparameter $\beta$ specifies the degree of robustness.
\subsection{Quantile regression for diffusion models for regression tasks}

Diffusion probabilistic models \cite{ho2020denoising} are primarily composed of two essential processes: a forward process that gradually adds Gaussian noise to a data sample, and a reverse process that transforms Gaussian noise to an empirical data distribution through a gradual denoising process. Conditional diffusion models \cite{saharia2022image} incorporate input samples to condition the denoising process. Image translation problems can be modeled as conditional diffusion models, represented as: \(p(y|x)\), where \(y\) is the target image and \(x\) is the input conditioning image. In this paper, we deal with image translation problems where the input images \(x\) are T1-weighted brain MRI images and the targets \(y\) are the corresponding T2-weighted images. The diffusion model \(f_{\theta}(x)\) is trained to recover T2-weighted images \(y\) from Gaussian noise \(\epsilon \sim \mathcal{N}(0,I)\) conditioned on the input T1-weighted images \(x\). For the details of the diffusion and conditional diffusion models, we refer to multiple works that provide a complete treatment of the mathematical formulations \cite{nichol2021improved, sohl2015deep, saharia2022image, ho2020denoising, saharia2022palette}. Instead of minimizing the mean squared error loss between the targets \(y\) and the estimates \(f_{\theta}(x)\) that yields a mean regression problem, the minimization problem in (\ref{q_loss}) is adopted to predict the \(\alpha\) quantiles of the target images. We show that in the presence of outliers in the training set, replacing the loss function in (\ref{q_loss}) with the proposed loss function in (\ref{robust_loss}) yields a model that is minimally affected by the outliers, making it closer to a model trained only using inlier samples. The details of the conducted experiments are presented in the following section.


\section{Experiments and results}
In this section, we evaluate our proposed method on a simple real dataset, a simulation-based dataset, and a medical image translation problem.
\subsection{Star cluster CYB OB1}
First, we start with a simple dataset on the star cluster CYB OB1 which was analyzed in \cite{neykov2012least}. This dataset consists of 47 observations from which four points with high leverage do not follow the trend of the rest of the data. It has one explanatory variable which is the logarithm of the effective temperature at the surface of stars. The independent variable is the logarithm of its light intensity.  The authors have shown the efficacy of least trimmed quantile regression compared to quantile regression using linear programming optimization to find the model's parameters. However, our goal is to investigate robustness in neural networks where the solution will be calculated using stochastic gradient descent (SGD). We estimate 0.25, 0.5, and 0.75 quantiles with a neural network.

We implement the linear quantile regression problem with a one-layer neural network with linear activation. Then we applied the three suggested robust methods TQR, RCP, and $\beta$-QR. We used GD with the ADAM optimizer to train the network. We chose the hyperparameters for each model (trimming percentage, L and $\beta$) using a grid search. We used a batch size of 47 and performed 5000 iterations. 

\begin{figure}[!tb] 
  \centering
  \includegraphics[width=0.9\columnwidth]{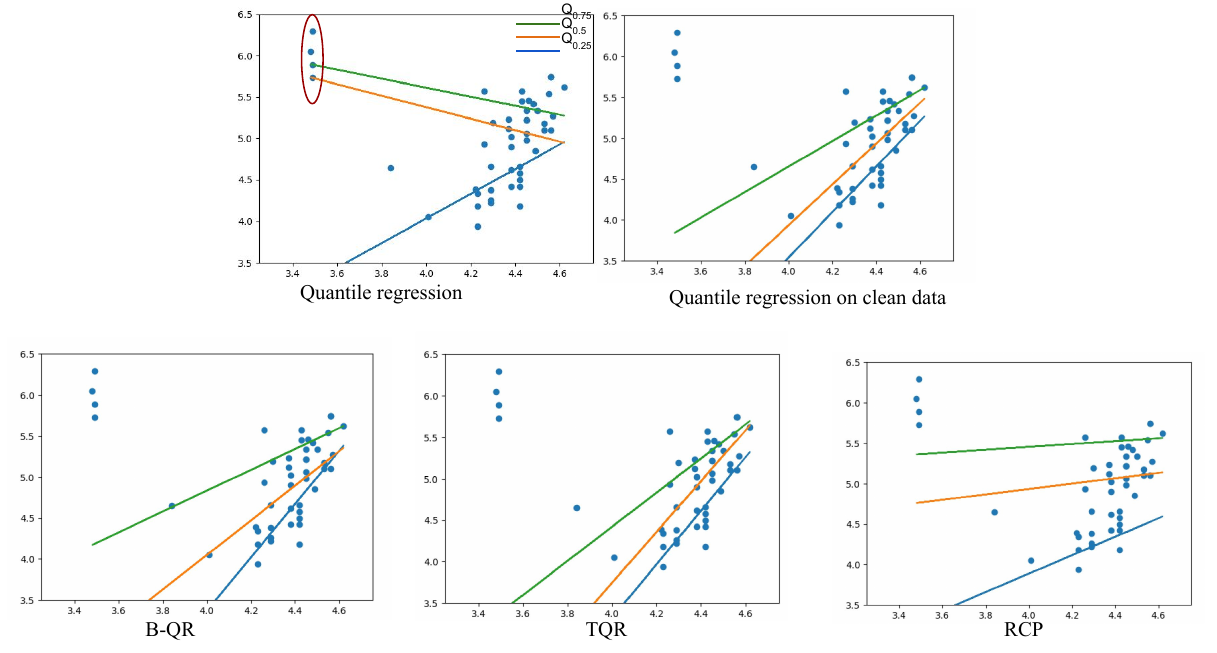} 
  \caption{Robust linear quantile regression using TQR, RCP, $\beta$ -QR for star cluster CYB OB1 dataset. 
  }
  \label{fig:linear}
  \vspace{-30pt}
\end{figure}

The results are shown in Fig. \ref{fig:linear}. For a quantitative comparison of the models We calculated the Frobenius norm between each estimated quantile and the solution, which was learned only using the inliers (Table \ref{tab:1} and Fig. \ref{fig:linear}). The $\beta$-QR method shows the best performance among the methods. For optimizing RCP cost, we used the Alternating Direction Method of Multipliers (ADMM) in which we split the objective into $\sum_{i}\rho_{\alpha}(y_{i}-f_{\theta}(x_{i})-\gamma_{i})$ and $\lambda \sum_{i} |\gamma_{i}|$. We optimized the former using GD with the ADAM optimizer, and for the latter, we used a proximal method for the $L_{1}$ objective:

\begin{align}
proxy_{\lambda,l_{1}}(x_{i}):=\begin{cases}
            x_{i}-\lambda   &\text{if $x_{i} > \lambda$ }  \\
            x_{i}+\lambda   &\text{if $x_{i} < \lambda$ }  \\
            0   &\text{otherwise.}
        \end{cases}
\end{align}

We iterated between optimization of the two components of the cost function until convergence. 

\subsection{Toy example for uncertainty estimation}
\begin{figure}[!htb]
\vspace{-10pt}
\centering
  \includegraphics[width=0.9\columnwidth]{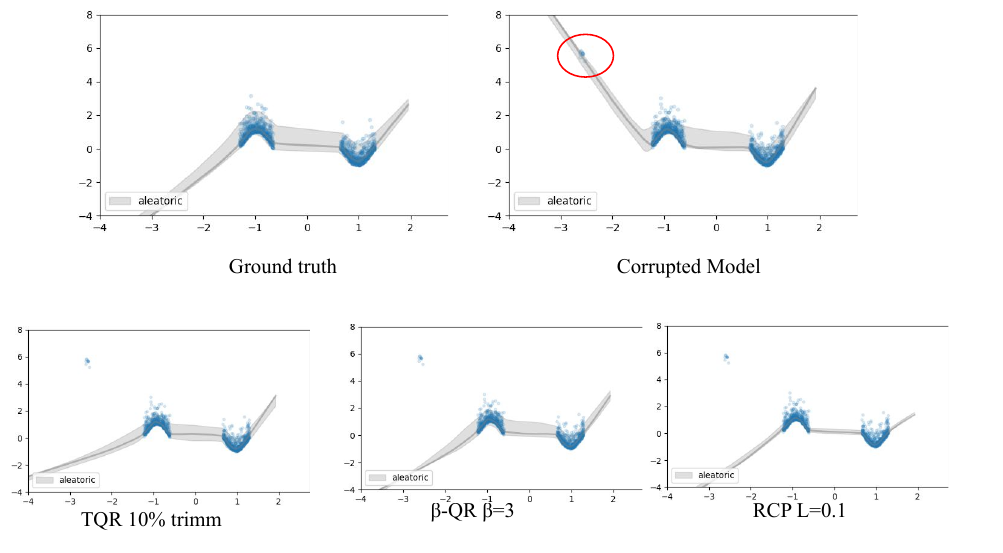}
  \vspace{-10pt}
  \caption{Robust non-linear quantile regression using TQR, RCP, $\beta$-QR using a simple neural network for a toy example. 
  }
  \label{fig:non-linear}
\vspace{-20pt}
\end{figure}

Here we used the simple synthetic dataset introduced in \cite{tagasovska2019single} to which we added 1\% outliers. Tagasovska and Lopez-Paz \cite{tagasovska2019single} applied simultaneous Quantile Regression (SQR) to estimate aleatoric uncertainty and suggested estimating all the quantile levels simultaneously.
We modeled the data using a three-layer neural network with ReLU activation function. We then applied the three robust methods TQR, RCP, and $\beta$-QR. We used SGD with the ADAM optimizer for training. We trained TQR and $\beta$-QR with a batch size of 128 and ran each for 500 epochs. RCP was trained for ten epochs and 500 steps of iterative optimization. We estimated the performance of the robust model for 0.25, 0.5 and 0.75 quantiles. Our results shown that $\beta$-QR estimates robust quantiles and comparable results to TQR (Fig. \ref{fig:non-linear}). 

\begin{figure*}[!bt]
\centering
\vspace{-10pt}
  \includegraphics[width=1.8\columnwidth]{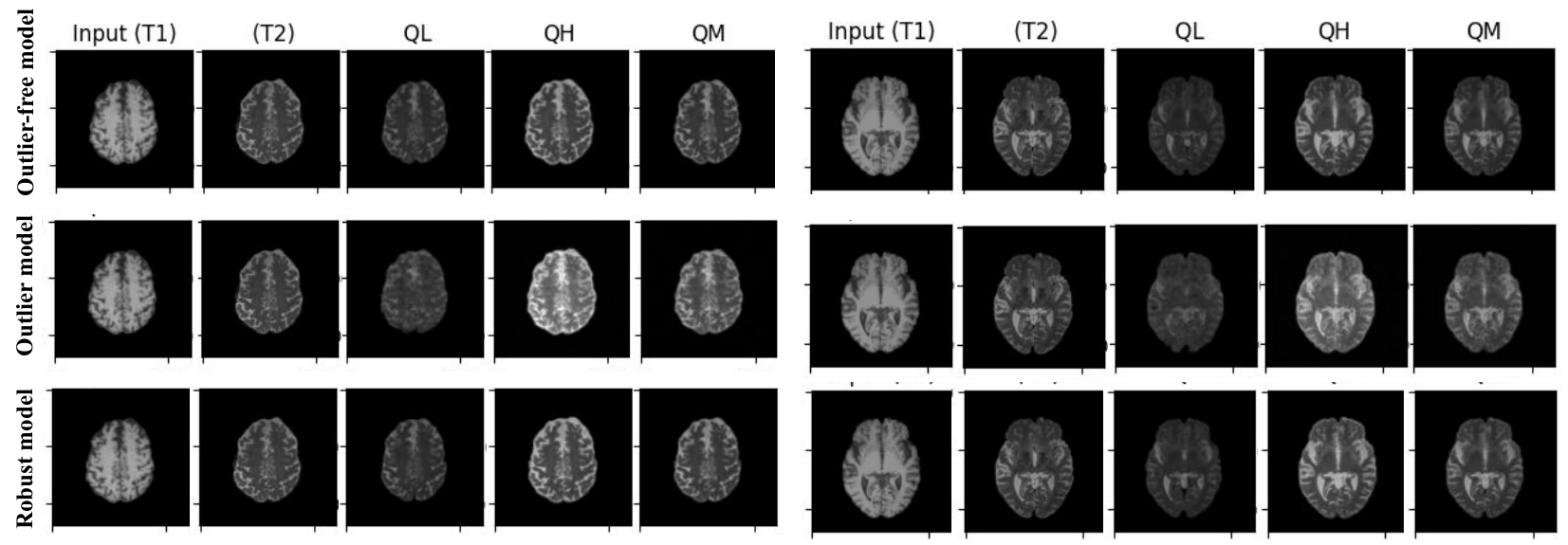}
  \vspace{-12pt}
  \caption{Estimating T2 MRI QL(0.05),QH(0.95),QM(0.5) for diffusion models from T1 MRI. Comparing the estimated quantiles using the non-robust and robust ($\beta$-QR) model with the outlier free model.}
\vspace{-15pt}
  \label{fig:brain}
\end{figure*}

\begin{table}[tb]
\vspace{-4pt}
\caption{Comparison of performance of TQR, RCP and $\beta$-QR. Each entry shows the Frobenius norm of the difference between the estimated quantiles and their (outlier-free) ground truth for the star cluster CYB OB1 dataset.   }
\centering
\begin{tabular}{ |c|c|c|c|} 
\hline
Method&	CYB-Q1&	CYB-Q2&	CYB-Q3\\
\hline
TQR & 1.04&	1.12&	1.12 \\ 
\hline
RCP &3.43&	4.82&	3.74 	\\
\hline
$\beta$-QR & 	0.93 &	0.77 &	0.85	\\
\hline
\end{tabular}
\label{tab:1}
\vspace{-10pt}
\end{table}

\subsection{Quantile regression for uncertainty estimation in diffusion models}

In this section, we present an experiment aimed at showcasing the effectiveness of our proposed robust quantile regression approach in a medical imaging task. Specifically, we focus on addressing the outlier problem in an image translation task, where we employ a diffusion model to predict various quantiles of T2-weighted brain MRI images based on input T1-weighted images.

Our training dataset consists of two distinct groups of subjects: (i) Lesion-free subjects that come from the Cam-CAN dataset (in-liers) \cite{taylor2017cambridge}, representing individuals without any brain lesions, and (ii) Lesion subjects that are sourced from the BRATS dataset \cite{menze2014multimodal} who do have brain lesions, thereby introducing outliers into the dataset. Training the diffusion model solely on the Cam-CAN data and using the loss function in (\ref{q_loss}) yields a reliable model that successfully captures the relationship between T1 and the quantiles of T2 images. However, introducing the ``outlier'' lesioned brain images from the BRATS dataset into the training set and using the same loss function significantly perturbs the training process, resulting in a notably less reliable model and corrupted quantiles. To mitigate the adverse effects of the outlier samples and restore model reliability, we integrate the proposed robust loss function presented in  (\ref{robust_loss}) into the training process. This loss function is designed to down-weight the influence of outliers during training, effectively bringing the model's performance closer to that of the model trained solely on clean data from the Cam-CAN dataset. The robust loss function was employed to train the model with the combined Cam-CAN and BRATS data sets. The results of this experiment are illustrated in Fig. \ref{fig:brain}, providing a qualitative comparison of the trained models.  Table \ref{tab:2} shows our quantitative results. These results unequivocally demonstrate that the inclusion of the robust loss function during model training significantly enhances the model's robustness to outliers, resulting in a reliable model that closely approximates the performance of the model trained exclusively on clean data. We estimated 0.05,0.5 and 0.95 for this dataset. For comparison of the robust and non-robust models, we calculated: (1) the MSE between the estimated quantiles and the outlier-free model predicted quantiles; and (2) the MSE between the predicted median and the ground-truth T2 image. We tuned the $\beta$ parameter using a validation set.

\section{Conclusion}

In this paper we introduced a robust quantile regression approach designed to enhance the reliability of deep learning models in the presence of outliers. Our method leverages concepts from robust divergences to down-weight outlier influence during training. We demonstrated the effectiveness of our approach on a simple yet real dataset, showcasing its ability to improve quantile regression accuracy compared to existing robust quantile regression methods. Extending the application to medical imaging, and demonstrating its practical utility, the proposed approach proved effective in mitigating outlier effects on training a diffusion model to translate MRI brain images from a T1-weighted to T2-weighted modality, bringing the performance closer to that of the model trained solely on clean data. The presented findings highlight the practical value of the proposed method, particularly in training scenarios compromised by outliers.

\begin{table}[tb]
\vspace{-4pt}
\caption{Comparing performance of $\beta$-QR with outlier free on baseline model. For the prediction error, MSE calculated between ground truth T2 and the median of each model }
\centering
\begin{tabular}{ |c|c|c|} 
\hline
Method&	Prediction error&	Quantile error\\
\hline
Outlier free & 0.0086&	- \\ 
\hline
Baseline &0.0132&0.0097	 	\\
\hline
$\beta$-QR & 0.0074&0.0013 	\\
\hline
TQR & 0.0107&0.0015	\\
\hline
\end{tabular}
\label{tab:2}
\end{table}

\bibliographystyle{ieee}
{\footnotesize
\bibliography{refs}}

\end{document}